\PassOptionsToPackage{hidelinks}{hyperref}
\PassOptionsToPackage{table}{xcolor}
\documentclass[sigconf, screen]{acmart}

\AtBeginDocument{%
  }

\setcopyright{acmlicensed}
\copyrightyear{2018}
\acmYear{2018}
\acmDOI{XXXXXXX.XXXXXXX}
\acmConference[arXiv Preprint]{}{arXiv}{2025}
\acmISBN{978-1-4503-XXXX-X/2018/06}

\acmSubmissionID{3972}

\usepackage{multirow}        
\usepackage{booktabs}        
\usepackage[capitalize,nameinlink]{cleveref} 
\begin{document}

\title{Mixture of Global and Local Experts with Diffusion Transformer for Controllable Face Generation}

\author{%
{\fontsize{10}{16}\selectfont
Xuechao Zou$^{1*}$,
Shun Zhang$^{1*}$,
Xing Fu$^{2}$,
Yue Li$^{3}$,
Kai Li$^{4}$,
Yushe Cao$^{4}$,
Congyan Lang$^{1\dagger}$,
Pin Tao$^{4}$,
Junliang Xing$^{4\dagger}$ \\
$^{1}$Beijing Jiaotong University,
$^{2}$Ant Group,
$^{3}$Qinghai University,
$^{4}$Tsinghua University \\
$^{*}$Equal contribution. \quad $^{\dagger}$Corresponding authors
}}

\renewcommand{\shortauthors}{Zou et al.}
\renewcommand\footnotetextcopyrightpermission[1]{} 
\settopmatter{printacmref=false} 


\begin{abstract}

Controllable face generation poses critical challenges in generative modeling due to the intricate balance required between semantic controllability and photorealism. While existing approaches struggle with disentangling semantic controls from generation pipelines, we revisit the architectural potential of Diffusion Transformers (DiTs) through the lens of expert specialization. This paper introduces Face-MoGLE, a novel framework featuring: (1) Semantic-decoupled latent modeling through mask-conditioned space factorization, enabling precise attribute manipulation; (2) A mixture of global and local experts that captures holistic structure and region-level semantics for fine-grained controllability; (3) A dynamic gating network producing time-dependent coefficients that evolve with diffusion steps and spatial locations. Face-MoGLE provides a powerful and flexible solution for high-quality, controllable face generation, with strong potential in generative modeling and security applications. Extensive experiments demonstrate its effectiveness in multimodal and monomodal face generation settings and its robust zero-shot generalization capability. Project page is available at \url{https://github.com/XavierJiezou/Face-MoGLE}.
\end{abstract}



\begin{teaserfigure}
  \includegraphics[width=\textwidth]{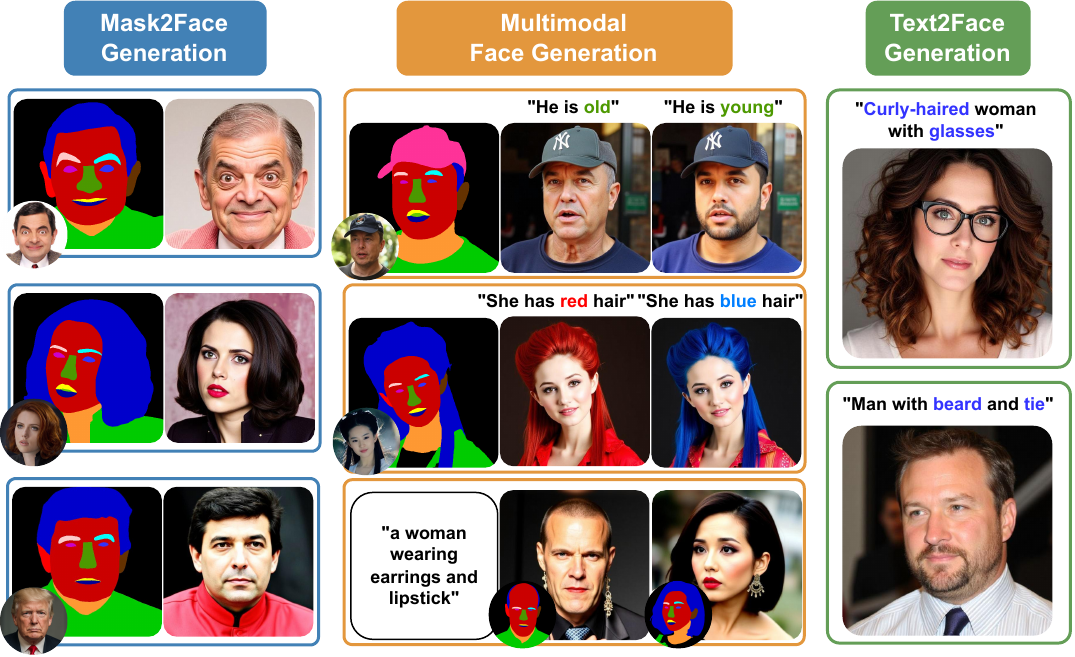}
  \caption{\textit{Mixture of Global and Local Experts with Diffusion Transformer} (Face-MoGLE) is a unified and flexible framework for high-quality and controllable face generation. It supports text-to-face synthesis (left), mask-to-face synthesis (right), and multimodal face generation guided jointly by text and masks (middle). By harmonizing global context modeling with local detail refinement, Face-MoGLE produces highly photorealistic results with enhanced semantic consistency and visual fidelity.}
  \label{fig: teaser}
\end{teaserfigure}


\maketitle

\section{Introduction}

    
Face generation has become a central task in computer vision, with wide-ranging applications in digital content creation~\cite{stylegan, styleclip}, virtual reality~\cite{avatarclip}, and human-computer interaction~\cite{wav2lip}. Beyond entertainment, this technology holds significant promise in security and public welfare. For example, it can assist criminal investigations by synthesizing suspect portraits from forensic sketches or textual descriptions~\cite{sketchgan}, and support the search for missing persons by reconstructing plausible appearances from partial visual cues. Generating realistic and contextually appropriate faces offers profound potential for numerous practical uses.

Face generation involves the core challenges of achieving both high image fidelity and controllability over various facial attributes. Controllable face generation, in particular, focuses on generating facial images that can manipulate specific attributes, such as identity, expression, or appearance. Early efforts in controllable face generation were dominated by generative adversarial networks (GANs)~\cite{stylegan, pggan, ganspace}, which enable high-resolution synthesis but often suffer from issues such as mode collapse, training instability, and limited adaptability to complex or multi-modal conditions. Flow-based methods~\cite{glow} offer invertibility and likelihood estimation but fall short in sample quality. Recently, diffusion models~\cite{ddpm, ddim, ldm} have become the de facto standard for high-fidelity image synthesis due to their strong generative performance and compatibility with conditional guidance, making them particularly effective for controllable face generation. Recently, several works~\cite{controlnet, instructpix2pix, t2i-adapter, pnp, ip-adapter, hico} have explored fine-tuning pre-trained diffusion models to support conditional inputs such as sketches or semantic maps, further enhancing controllability.

Despite the success of diffusion models, most state-of-the-art models rely on U-Net backbones~\cite{ldm}, which suffer from two inherent limitations. First, the convolutional inductive bias of U-Nets restricts their ability to model long-range dependencies essential for holistic facial consistency, while their entangled feature representations conflate structural and textural information, hindering precise attribute control. Second, existing methods~\cite{collaborative, uniteandconquer} tightly couple semantic masks with generation by directly concatenating masks and latent codes, which propagates mask errors to texture synthesis and limits fine-grained control over local regions. This coupling also imposes impractical requirements for pixel-perfect masks during inference, compromising zero-shot generalization. These dual limitations motivate our architectural redesign to strengthen global modeling while decoupling semantic guidance from low-level synthesis. While pre-trained foundational diffusion transformers (DiT)~\cite{dit, sd3, flux.1-dev} have recently demonstrated stronger generalization and higher fidelity compared to U-Net-based diffusion models, their application to face generation—especially under complex, multi-modal conditional settings—remains largely underexplored.

To address the limitations of existing controllable face generation methods, we propose Face-MoGLE, a novel framework that decouples semantic masks into independent binary components, each corresponding to a distinct facial attribute such as hair, face contour, or nose. This semantic decoupling enables precise, region-specific control and lays the foundation for targeted refinement. To effectively model both global structure and local detail, we introduce a Mixture of Experts (MoE) design: global experts capture holistic relationships across facial regions (e.g., ensuring spatial alignment between hair and face), while local experts focus on fine-grained features within individual regions (e.g., refining the texture of eyebrows or hair strands). These experts are seamlessly integrated into the Diffusion Transformer (DiT) backbone, which provides a powerful foundation for high-fidelity image synthesis and temporal-aware conditioning throughout the denoising process. Together, these components enable Face-MoGLE to achieve both semantically controllable and visually realistic face generation. In summary, our main contributions are as follows:

\begin{itemize}
    \item We propose a unified and modular generation framework based on the Diffusion Transformer (DiT), which decouples semantic masks into binary components to enable structured and disentangled condition modeling.
    \item We design a Mixture of Global and Local Experts (MoGLE) architecture, where global experts capture holistic facial structures and local experts refine region-specific details for improved semantic alignment and visual fidelity.
    \item We introduce a diffusion-aware dynamic gating network that adaptively blends expert outputs with spatial and temporal awareness, enabling fine-grained control throughout the denoising process.
\end{itemize}

Experimental results showed that Face-MoGLE significantly outperformed state-of-the-art (SOTA) controllable face generation models across multiple benchmarks. Compared to strong diffusion baselines such as PixelFace+ \cite{pixelfaceplus} and DDGI~\cite{ddgi}, our model achieved better FID scores and higher condition consistency. 


We extend the FFHQ-Text~\cite{ffhq_text} dataset with high-quality semantic segmentation masks to enable precise region-level control for multimodal input tasks. Testing on this expanded dataset confirms Face-MoGLE’s robust zero-shot generalization. Notably, Face-MoGLE produces images with high perceptual realism that can evade SOTA face forgery detectors, underscoring its promise in generative and security applications. As illustrated in \cref{fig: teaser}, even without explicit single-modal training, it delivers strong results in Mask2Face and Text2Face tasks without retraining or architectural changes. Overall, by leveraging architectural innovations and explicit condition disentanglement, Face-MoGLE offers a powerful and flexible solution for high-quality, controllable face generation.

\section{Related Work}

\subsection{Diffusion Model}

Diffusion models~\cite{ddpm, improved-ddpm, guided-diffusion} have emerged as a powerful class of generative models, demonstrating superior performance to GANs in image synthesis, which had long been dominated by GAN-based approaches~\cite{maskgan, tedigan, lggan, ffhq}. Inspired by the physical diffusion process~\cite{thermodynamics}, these models learn to generate data by reversing a gradual noising process. In the forward pass, data is progressively corrupted with Gaussian noise over multiple timesteps, while the model is trained to recover the original sample through a denoising process.

The denoising diffusion probabilistic model proposed by Ho et al.~\cite{ddpm} laid the foundation for this approach with impressive image synthesis performance. Later, Nichol and Dhariwal proposed the classifier-free guidance method ~\cite{cfg}, which enabled conditional generation without relying on external classifiers. Recent advancements have transitioned from modeling in pixel space to latent space. LDM~\cite{ldm} employs a U-Net~\cite{unet} architecture for efficient denoising in a compressed latent space. More recent works replace the convolutional U-Net with vision transformers~\cite{vit, dit}, leveraging their global attention mechanisms and geometry-aware positional encodings to capture spatial dependencies better. These Diffusion Transformers (DiTs)~\cite{dit, flux.1-dev, vasa-1, sd3} have demonstrated strong scalability, with performance improvements that correlate with model capacity and training compute, establishing them as the new state-of-the-art in diffusion-based generation tasks.

\subsection{Face Generation}

Face generation has become a cornerstone task in computer vision, with increasing demands for controllability and high-fidelity synthesis. Current approaches can be broadly categorized into GAN-based, diffusion-based, and hybrid methods, and are typically applied to unimodal (e.g., mask or text) or multimodal settings.

In unimodal face generation, mask-to-face and text-to-face are two representative tasks. For mask-to-face generation, methods like MaskGAN~\cite{maskgan} use semantic masks to enable interactive and diverse facial manipulation. INADE~\cite{inade} introduces stochastic sampling on class distributions to enhance diversity, while E2Style~\cite{e2style} focuses on efficient and accurate StyleGAN inversion. SemFlow~\cite{semflow} further unifies image synthesis and segmentation using rectified flow, achieving reversible transformations. In text-to-face generation, clip2latent~\cite{clip2latent} and GCDP~\cite{gcdp} employ CLIP-based guidance and semantic layout generation to improve text-image alignment. E3-FaceNet~\cite{e3facenet} additionally introduces 3D awareness and geometric regularization to improve realism and view consistency.

In multimodal face generation, the core challenge is aligning diverse conditions—such as text, masks, or sketches—while preserving image fidelity. TediGAN~\cite{tedigan} and PixelFace+~\cite{pixelfaceplus} enable flexible content creation using both textual and visual inputs. MM2Latent~\cite{mm2latent} directly maps multimodal signals to the GAN latent space for efficient generation. Diffusion-based methods like Collaborative Diffusion~\cite{collaborative} and UaC~\cite{uniteandconquer} support plug-and-play multimodal synthesis. DDGI~\cite{ddgi} integrates GAN inversion with diffusion features to handle multi-condition face generation.

While prior works have made notable progress in specific settings, they often struggle to balance fine-grained semantic control with high-quality synthesis, especially under zero-shot or compositional generalization. In contrast, our proposed Face-MoGLE achieves strong results across unimodal and multimodal tasks. 

\subsection{Mixture of Experts}

The Mixture of Experts (MoE) model is a neural architecture that enhances scalability and specialization by partitioning the input space among several expert networks. Each expert learns to model a subset of the data distribution, while a gating network dynamically assigns tokens to relevant experts based on input semantics. The concept was first introduced by Hinton et al.~\cite{hinton1991adaptive}, who proposed a supervised learning framework in which each expert specializes in a distinct region of the input space. This early work provided a theoretical link between modular neural networks and competitive learning, and laid the foundation for modern sparse expert systems.

Subsequent advances have integrated MoEs into deep learning. Notably, \citet{shazeer2017outrageously} introduced the Sparsely-Gated Mixture-of-Experts, demonstrating large-scale training efficiency. GShard~\cite{lepikhin2020gshard} and Switch Transformers~\cite{fedus2021switch} refined these ideas, improving training stability and enabling trillion-parameter models. In the vision domain, V-MoE~\cite{riquelme2021scaling} brought sparse expert routing into Vision Transformers~\cite{vit}. Inspired by these works, our Face-MoGLE leverages a mixture of global and local experts within a diffusion transformer. This enables high-fidelity, controllable face generation by dynamically selecting experts throughout the denoising process.

\section{Method}

\subsection{Overall Pipeline}

We build upon the DiT~\cite{dit} architecture, utilizing FLUX~\cite{flux.1-dev} as our foundational model, to introduce \textit{Mixture of Global and Local Experts (MoGLE)}—a simple yet powerful framework for fine-grained controllable face generation. This design accepts multimodal control conditions, aiming to harmonize global context modeling with local detail refinement for semantic masks within a unified model.

\paragraph{\textbf{Training.}} \label{sec: training}

\begin{figure}[h]
    \centering
    \includegraphics[width=0.95\linewidth]{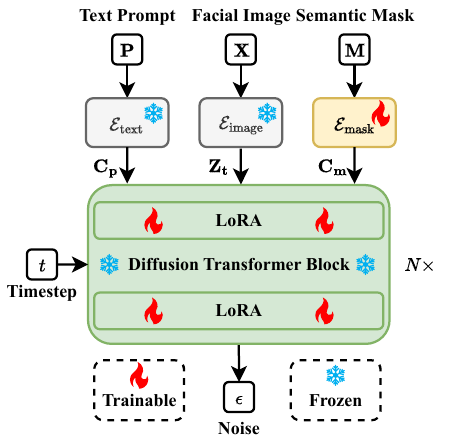}
    \caption{Training pipeline of the diffusion transformer.}
    \label{fig: pipeline}
\end{figure}

The training process is entirely conducted in the \textit{latent space}, where both the forward diffusion and reverse denoising occur. This design avoids direct modeling in pixel space, which significantly improves computational efficiency~\cite{ldm}. As illustrated in \cref{fig: pipeline}, the model is conditioned on multimodal signals, including a text prompt and a semantic mask, and is trained to predict the noise added to the latent image tokens during the diffusion process.

Let the input text prompt be \( \mathbf{P} \), the facial image be \( \mathbf{X} \in \mathbb{R}^{H \times W \times 3} \), and the semantic mask be \( \mathbf{M} \in \mathbb{R}^{H \times W \times 3} \). These inputs are transformed into token sequences via the following encoders:
\begin{equation}
\mathbf{C}_p = \mathcal{E}_{\text{text}}(\mathbf{P}), \quad 
\mathbf{Z} = \mathcal{E}_{\text{image}}(\mathbf{X}), \quad
\mathbf{C}_m = \mathcal{E}_{\text{mask}}(\mathbf{M}),
\end{equation}
where \( \mathbf{C}_p \in \mathbb{R}^{L' \times d'} \), \( \mathbf{Z} \in \mathbb{R}^{L \times d} \), and \( \mathbf{C}_m \in \mathbb{R}^{L \times d} \) denote the token sequences output by the text, image, and mask encoders, respectively. \( L' \) and \( L \) are the token lengths, and \( d' \), \( d \) are the embedding dimensions.

The text encoder \( \mathcal{E}_{\text{text}} \) jointly leverages CLIP~\cite{clip} and H5~\cite{h5}, combining strong generalization with stylistic richness. The image encoder \( \mathcal{E}_{\text{image}} \) is based on the encoder of a pretrained VAE~\cite{vae, vq-vae}, which maps the image from pixel space to latent token representations. Both encoders are kept frozen during training.

In contrast, the mask encoder \( \mathcal{E}_{\text{mask}} \) is a carefully designed component intended to enhance fine-grained and structured semantic control. We propose a \textit{Mixture of Global and Local Experts} (MoGLE) architecture to obtain rich and flexible token-level representations from semantic masks. This design addresses the lack of spatial controllability in pretrained text-to-image generation DiTs~\cite{ldm, flux.1-dev}. Details can be found in \cref{sec: mogle}.

At each training step, a timestep \( t \in \{1, \dots, T\} \) is sampled, and Gaussian noise is added to the image tokens:
\begin{equation}
\mathbf{Z}_t = \sqrt{\bar{\alpha}_t} \mathbf{Z} + \sqrt{1 - \bar{\alpha}_t} \boldsymbol{\epsilon}, \quad \boldsymbol{\epsilon} \sim \mathcal{N}(0, \mathbf{I}),
\end{equation}
where \( \bar{\alpha}_t \) denotes the cumulative product of the forward noise schedule. The resulting noisy tokens \( \mathbf{Z}_t \in \mathbb{R}^{L \times d} \) are fed into the denoising network. To improve robustness, we apply a drop probability of 0.1 independently to each condition (i.e., text prompt or semantic mask) during training, following previous works~\cite{ip-adapter, ominicontrol, vasa-1}. This allows the model to gracefully handle cases where one or both modalities are absent (i.e., set to \( \emptyset \)), thereby enabling flexible and controllable face generation.

Our denoising module is a diffusion transformer composed of \( N \) stacked blocks, each consisting of a frozen transformer backbone and trainable Low-Rank Adaptation (LoRA)~\cite{lora, ominicontrol} modules for efficient fine-tuning:
\begin{equation}
\hat{\boldsymbol{\epsilon}} = f_\theta\left( \mathbf{Z}_t, t, \mathbf{C}_p, \mathbf{C}_m \right),
\end{equation}
where \( f_\theta \) denotes the denoising network, with only the LoRA modules updated during training. The model is optimized by minimizing the mean squared error between the predicted and true noise:
\begin{equation}
\mathcal{L}_{\text{MSE}} = \mathbb{E}_{t, \mathbf{Z}, \boldsymbol{\epsilon}} \left[ \left\| \hat{\boldsymbol{\epsilon}} - \boldsymbol{\epsilon} \right\|_2^2 \right].
\end{equation}
This objective guides the model to iteratively denoise latent image tokens while attending to textual and semantic signals.

\paragraph{\textbf{Sampling.}} \label{sec: sampling}
During inference, we adopt an improved sampling procedure~\cite{sd3, flux.1-dev, ominicontrol} to iteratively denoise latent image tokens, initialized from pure Gaussian noise. The text prompt and semantic mask are encoded the same way as during training, using the frozen text encoder and the trained mask encoder, respectively. These condition tokens guide the denoising process toward generating semantically faithful and structurally aligned images. To support flexible generation, our model allows either condition to be dropped at test time by setting it to an empty input. This mirrors the condition drop strategy used during training and enables diverse use cases, such as semantic face synthesis without text, or text-to-face generation from text alone. After denoising completes, the generated latent tokens are decoded into the final image using the pretrained VAE~\cite{vae, vq-vae} decoder. The output image reflects high-level semantics and spatial structure derived from the conditioning inputs, enabling high-quality and controllable face generation.

\subsection{Mixture of Global and Local Experts} \label{sec: mogle}

\paragraph{\textbf{Global and Local Experts.}} \label{sec: gle}

\begin{figure*}[htbp]
    \centering
    \includegraphics[width=0.90\textwidth]{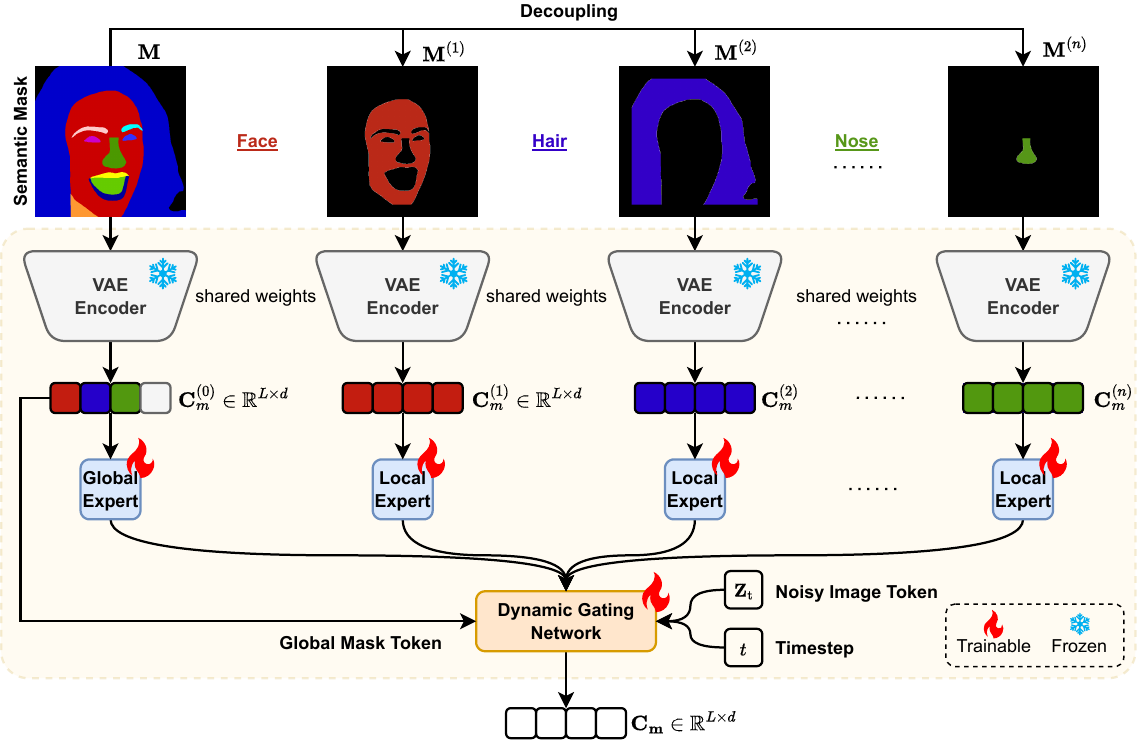}
    \caption{Architecture of the Mixture of Global and Local Experts (MoGLE) designed for semantic mask embedding.}
    \label{fig: mogle}
\end{figure*}

To obtain expressive and controllable semantic representations from the input mask, we design a Mixture of Global and Local Experts architecture, as shown in \cref{fig: mogle}. The motivation is twofold: (1) to extract global structural priors from the full-face layout; and (2) to model region-specific semantics for enhanced controllability and fidelity in face generation.

Given a semantic mask \( \mathbf{M} \in \mathbb{R}^{H \times W \times 3} \), we first decouple it into \( n \) binary masks \( \{ \mathbf{M}^{(i)} \}_{i=1}^n \), each representing a semantic region (e.g., face, hair, nose). All masks are passed through a shared frozen VAE encoder \( \mathcal{E}_{\text{VAE}} \)~\cite{vae, vq-vae}, producing a sequence of latent tokens:
\begin{equation}
\mathbf{C}_m^{(i)} = \mathcal{E}_{\text{VAE}}(\mathbf{M}^{(i)}) \in \mathbb{R}^{L \times d}, \quad i = 0, 1, \ldots, n
\end{equation}
where \( \mathbf{M}^{(0)} \) denotes the full mask, used to derive global context. Each token sequence \( \mathbf{C}_m^{(i)} \) is processed by its corresponding expert:
\begin{equation}
\mathbf{C}_m^{(i)\prime} = \text{Expert}_i(\mathbf{C}_m^{(i)}) \in \mathbb{R}^{L \times d}. \quad i = 0, 1, \ldots, n
\end{equation}

The global expert captures high-level spatial priors, while local experts focus on fine-grained, region-specific semantics. This cooperative modeling enables the system to maintain structural consistency while improving the fidelity of generated faces.

\paragraph{\textbf{Dynamic Gating Network.}} \label{sec: dgn}

\begin{figure}[htbp]
    \centering
    \includegraphics[width=0.95\linewidth]{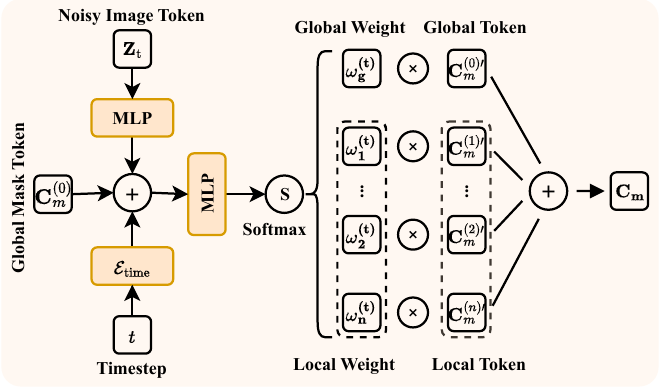}
    \caption{Structure of our dynamic gating network.}
    \label{fig: gating}
\end{figure}

To dynamically integrate expert outputs across the diffusion process, we introduce a diffusion-aware gating network \( g_\theta \), illustrated in \cref{fig: gating}. This module takes the current noisy latent tokens \( \mathbf{Z}_t \), a learned timestep embedder \( \mathcal{E}_{\text{time}}(t) \), and the global mask token \( \mathbf{C}_m^{(0)} \), and produces normalized weights:
\begin{equation}
\begin{aligned}
[\mathbf{\omega_g^{(t)}}, \mathbf{\omega_1^{(t)}}, \ldots, \mathbf{\omega_n^{(t)}}] = 
g_\theta(\mathbf{Z}_t, \mathcal{E}_{\text{time}}(t), \mathbf{C}_m^{(0)}) \\
\text{s.t.} \quad \mathbf{\omega_g^{(t)}} + \sum_{i=1}^{n} \mathbf{\omega_i^{(t)}} = 1, \quad \mathbf{\omega_g^{(t)}}, \mathbf{\omega_i^{(t)}} \in [0, 1]
\end{aligned}
\end{equation}
where \( \mathbf{\omega_g^{(t)}} \) and \( \mathbf{\omega_i^{(t)}} \) denote the spatial weight maps for the global expert and the \( i \)-th local expert at timestep \( t \), respectively. Unlike static fusion, our gating mechanism produces spatially varying weights that evolve during the denoising process. The final semantic embedding is computed as:
\begin{equation}
\mathbf{C}_m = \mathbf{\omega_g^{(t)}} \cdot \mathbf{C}_m^{(0)\prime} + \sum_{i=1}^{n} \mathbf{\omega_i^{(t)}} \cdot \mathbf{C}_m^{(i)\prime}.
\end{equation}

\begin{figure}[htbp]
    \centering
    \includegraphics[width=0.95\columnwidth]{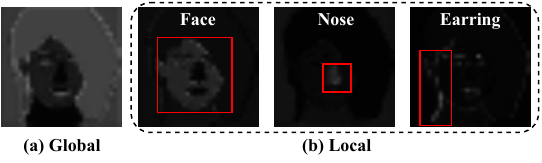}
    \caption{Visualization of global and partial local weight map.}
    \label{fig: weight_map}
\end{figure}

To better understand the gating behavior, we visualize the spatial weight maps predicted for both global and selected local experts in \cref{fig: weight_map}. These maps reveal that different semantic regions are adaptively emphasized at stages of the diffusion process, validating the gating network’s semantic awareness and spatial adaptivity.

\begin{table*}[htbp]
\centering
\caption{Comparison of face generation methods on the MM-CelebA-HQ dataset across different tasks.}
\label{tab: experiment_sota}
\renewcommand{\arraystretch}{1.2}  
\begin{tabular}{l c c ccc c c c}
\toprule
\multirow{2}{*}{\textbf{Method}} &
\multirow{2}{*}{\textbf{Venue}} &
\multirow{2}{*}{\textbf{Paradigm}} &
\multicolumn{3}{c}{\textbf{Image Generation Quality}} &
\multicolumn{2}{c}{\textbf{Condition Alignment}} &
\multirow{2}{*}{\textbf{IR~$\uparrow$}} \\
\cmidrule(lr){4-6}
\cmidrule(lr){7-8}
& & & FID~$\downarrow$ & KID~$\downarrow$ & CMMD~$\downarrow$ & Mask~$\downarrow$ & Text~$\uparrow$ & \\
\midrule
\multicolumn{9}{c}{\textbf{Multimodal Face Generation}} \\
\midrule
TediGAN~\cite{tedigan}
  & CVPR-21
  & GAN
  & 83.35 & 74.75 & 1.562 & \underline{2.46} & 23.90 & -0.1446 \\

UaC~\cite{uniteandconquer}
  & CVPR-23
  & Diffusion
  & 75.35 & 63.78 & 1.982 & 3.41 & 25.52 & -0.4001 \\

Collaborative~\cite{collaborative}
  & CVPR-23
  & Diffusion
  & \underline{24.48} & \underline{13.50} & \underline{0.734} & 3.22 & 24.51 & -0.1265 \\

PixelFace+~\cite{pixelfaceplus}
  & ACM MM-23
  & GAN
  & 65.53 & 53.90 & 1.273 & 2.61 & \underline{26.16} & \underline{0.6403} \\

DDGI~\cite{ddgi}
  & CVPR-24
  & GAN \& Diffusion
  & 46.68 & - & - & - & - & - \\

\rowcolor{gray!20}Face-MoGLE (Ours)
  & -
  & Diffusion
  & \textbf{22.24} & \textbf{10.87} & \textbf{0.477} & \textbf{2.44} & \textbf{26.32} & \textbf{0.7014} \\
\midrule
\multicolumn{9}{c}{\textbf{Mask-to-Face Generation}} \\
\midrule
INADE~\cite{inade}
  & CVPR-21  
  & GAN
  & \underline{21.09} & \underline{11.24} & 1.871 & 2.57 & \underline{24.85} & -0.0234 \\

E2Style~\cite{e2style}
  & TIP-22
  & GAN 
  & 38.44 & 21.22 & \underline{1.129} & \underline{2.36} & 24.75 & \underline{0.1649} \\

SemFlow~\cite{semflow}
  & NeurIPS-24  
  & Flow 
  & 56.65 & 41.48 & 1.767 & \textbf{2.30} & \textbf{25.65} & -0.004 \\

\rowcolor{gray!20}Face-MoGLE (Ours)
  & -
  & Diffusion
  & \textbf{19.63} & \textbf{8.29} & \textbf{0.399} & 2.57 & 24.53 & \textbf{0.0398} \\
\midrule
\multicolumn{9}{c}{\textbf{Text-to-Face Generation}} \\
\midrule
clip2latent~\cite{clip2latent}
  & BMVC-22
  & GAN
  & \underline{63.89} & \underline{38.55} & \underline{1.410} & 5.05 & \underline{27.56} & \textbf{1.0129} \\

GCDP~\cite{gcdp}
  & ICCV-23
  & Diffusion
  & 72.67 & 43.81 & 1.456 & \underline{4.68} & 25.94 & 0.4563 \\

E3-FaceNet~\cite{e3facenet}
  & ICML-24
  & GAN
  & 70.89 & 47.82 & 2.749 & \textbf{4.64} & \textbf{27.94} & 0.8150 \\

\rowcolor{gray!20}Face-MoGLE (Ours)
  & -
  & Diffusion
  & \textbf{34.81} & \textbf{21.85} & \textbf{0.636} & 4.94 & 26.91 & \underline{0.9527} \\
\bottomrule
\end{tabular}
\end{table*}

\begin{figure*}[htbp]
    \centering
    \includegraphics[width=0.95\textwidth]{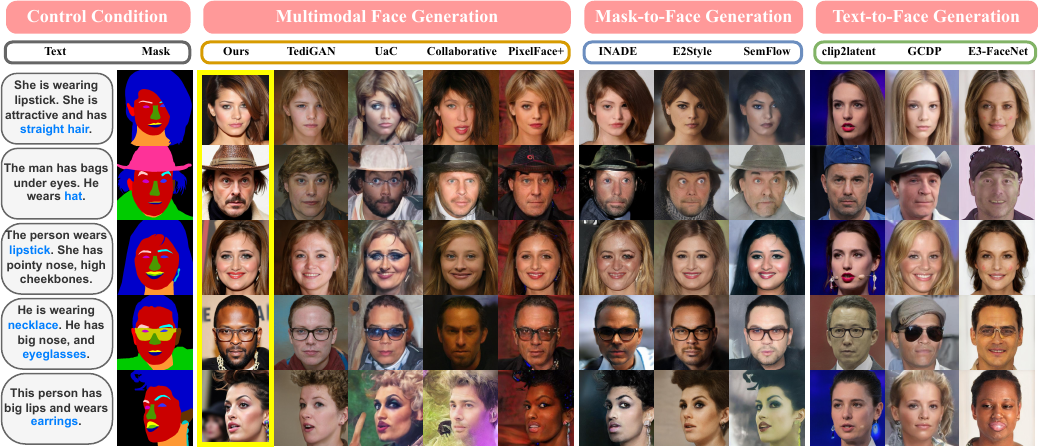}
    \caption{Visualization results from different methods. The figure compares three generation paradigms: multimodal face synthesis (left) and two unimodal tasks — mask-to-face and text-to-face generation (middle and right). From top to bottom, each row highlights attribute-specific synthesis: straight hair, hat, lipstick, necklace, and earrings.  Our method demonstrates superior alignment with the textual descriptions, semantic mask, and higher visual fidelity compared to other baseline models.}
    \label{fig: vis_results}
\end{figure*}

\section{Experiments}
\subsection{Datasets}
In this study, we employ two benchmark datasets: MM-CelebA-HQ~\cite{tedigan}, an enriched version of CelebAMask-HQ~\cite{maskgan} featuring high-resolution facial images annotated with attributes, utilized for both training and evaluation; and FFHQ-Text~\cite{ffhq_text}, a meticulously curated subset of FFHQ~\cite{ffhq}, comprising high-quality images of female faces accompanied by nuanced textual descriptions. We further refine this dataset into a multimodal variant, referred to as MM-FFHQ-Female, designed explicitly for zero-shot evaluation. In accordance with~\cite{collaborative}, the initial 27,000 pairs from MM-CelebA-HQ are allocated for training, while the remaining 3,000 serve as the test set. To produce semantic masks for FFHQ-Text, we leverage two pretrained facial parsing models—FaRL~\cite{farl} and SegFace~\cite{segface}. For masks with overall accuracy (OA) below 0.8, we conduct manual annotation, whereas for those achieving OA $\geq$ 0.8, we adopt a randomized sampling strategy, comprising 90\% from FaRL and 10\% from SegFace. More details of datasets are available in~\cref{sec: dataset details}.

\subsection{Implementation Details}
Our framework builds upon the open-source FLUX.1-dev~\cite{flux.1-dev}. During training, we set the batch size to 8 and utilize mixed-precision with the brain floating-point format to enhance computational efficiency and reduce memory usage. The optimizer employed is Prodigy~\cite{prodigy}, with a base learning rate of 1.0 and a weight decay of 0.01. A dropout probability 0.1 is applied to each controllable condition, excluding the global mask. The Low-Rank Adaptation (LoRA)~\cite{lora, ominicontrol} is configured with a rank of 4 and a scaling factor of 4. A fixed random seed of 42 is used for all experiments to ensure reproducibility. Inference is performed with 28 sampling steps, and training is conducted for 4,000 steps, taking approximately 12 hours on a workstation equipped with 8$\times$NVIDIA A100 80G GPUs.

\begin{table*}[htbp]
\centering
\caption{Comparison of multimodal face generation methods on the MM-FFHQ-Female dataset under zero-shot setting.}
\label{tab: experiment_zero_sota}
\renewcommand{\arraystretch}{1.2}
\begin{tabular}{l c c ccc c c c}
\toprule
\multirow{2}{*}{\textbf{Method}} &
\multirow{2}{*}{\textbf{Venue}} &
\multirow{2}{*}{\textbf{Paradigm}} &
\multicolumn{3}{c}{\textbf{Image Generation Quality}} &
\multicolumn{2}{c}{\textbf{Condition Alignment}} &
\multirow{2}{*}{\textbf{IR~$\uparrow$}} \\
\cmidrule(lr){4-6}
\cmidrule(lr){7-8}
\textbf{} & \textbf{} & \textbf{} & \textbf{FID~$\downarrow$} & \textbf{KID~$\downarrow$} & \textbf{CMMD~$\downarrow$} & \textbf{Mask~$\downarrow$} & \textbf{Text~$\uparrow$} & \textbf{} \\
\midrule
TediGAN~\cite{tedigan} & CVPR-21 & GAN & 122.47 & 92.72 & \textbf{1.091} & 3.32 & 25.03 & -0.4847 \\
UaC~\cite{uniteandconquer} & CVPR-23 & Diffusion & 86.58 & 46.57 & 1.796 & 3.11 & \underline{26.97} & -0.8851 \\
Collaborative~\cite{collaborative} & CVPR-23 & Diffusion & 93.31 & 62.84 & 2.178 & 3.85 & 23.03 & -1.2101 \\
PixelFace+~\cite{pixelfaceplus} & ACM MM-23 & GAN & \underline{76.45} & \underline{38.95} & 1.917 & \underline{3.08} & 26.60 & \underline{-0.2616} \\
\rowcolor{gray!20}Face-MoGLE (Ours) & - & Diffusion & \textbf{62.93} & \textbf{31.27} & \underline{1.238} & \textbf{2.77} & \textbf{28.06} & \textbf{0.1801} \\
\bottomrule
\end{tabular}
\end{table*}

\subsection{Evaluation Metrics}


\paragraph{\textbf{Image Quality}  }
The visual fidelity of generated images is a crucial indicator of model performance. We employ the following widely adopted metrics:  
\textit{Fréchet Inception Distance (FID)} \cite{fid} quantifies the distributional discrepancy between generated and real images in the Inception feature space, with lower values indicating higher quality.  
\textit{Kernel Inception Distance (KID)} \cite{kid}, similar in spirit to FID, relies on kernel-based methods to provide an unbiased and more stable estimate. For readability, we multiply the KID by 1,000.  
\textit{CLIP Maximum Mean Discrepancy (CMMD)} \cite{cmmd} measures the alignment of conditional distributions, making it particularly suitable for conditional generation tasks.  

\paragraph{\textbf{Text Consistency}}  
To evaluate how well the generated images align semantically with the input textual descriptions, we utilize the \textit{CLIP Score} \cite{clip}. This metric leverages the CLIP vision-language model to compute similarity between text and image embeddings, with higher scores indicating more substantial semantic alignment. For better readability, we report the scores in percentage format.

\paragraph{\textbf{Mask Consistency}}  
We use \textit{DINO Structure Distance} to measure how well the generated image aligns with source images under the guidance of the input semantic mask, which compares the self-similarity matrices of features from the DINO-ViT~\cite{dino}. A smaller value indicates higher mask (structural) consistency.


\paragraph{\textbf{Human Preference}}  
To reflect subjective human perception, we incorporate the \textit{Image Reward} (IR) score \cite{image-reward}, derived from a learned aesthetic scoring model. This metric estimates the visual appeal and coherence of generated images and has been shown to correlate closely with human judgments.

\paragraph{\textbf{Deepfake Detection}}
To further assess the realism of our synthesized facial images, we evaluate their ability to evade deepfake detection. It is measured using Area Under the Receiver Operating Characteristic Curve (AUC), Equal Error Rate (EER), and Average Precision (AP)~\cite{npr, wavelet-clip}. AUC and EER values close to 0.5 indicate increased similarity to authentic images and detector confusion, while lower AP reflects reduced confidence in fake identification. 

\subsection{Comparison with State-of-the-Art Methods}

To evaluate the effectiveness of Face-MoGLE, we conduct comprehensive comparisons with representative state-of-the-art methods across three face generation tasks. As shown in \cref{tab: experiment_sota}, our model consistently performs well across key metrics of generated image quality, condition alignment, and human perceptual preference.

\subsubsection{Multimodal Face Generation}

Face-MoGLE achieves the best results in FID (22.24), KID (10.87), CMMD (0.477), mask alignment (2.44), and also ranks first in text alignment (26.32), indicating strong generation fidelity and semantic controllability. Since DDGI~\cite{ddgi} has not released its code, we directly copy the results reported in its paper. Compared to the strongest diffusion-based baseline Collaborative~\cite{collaborative}, our method brings consistent improvements across all metrics, validating the effectiveness of the proposed framework.

\subsubsection{Monomodal Face Generation}


\paragraph{Mask2Face Generation}

For mask-to-face generation, Face-MoGLE again achieves the best results in FID (19.63), KID (8.29), and CMMD (0.399). The mask alignment score (2.57) matches top-performing methods, demonstrating the model's strength in preserving spatial structure while generating perceptually compelling faces.

\paragraph{Text2Face Generation}

In the text-to-face setting, Face-MoGLE achieves superior performance in terms of generation fidelity and condition consistency. Specifically, it obtains the best results across all image quality metrics, with an FID of 34.81, KID of 21.85, and CMMD of 0.636. Compared to other methods, Face-MoGLE is more effective at generating realistic and condition-consistent facial images from text descriptions, highlighting the effectiveness of our diffusion-based approach.

\subsubsection{Visualization and Human Preference}


We use the IR score~\cite{image-reward} to assess the perceptual quality of the generated faces. As shown in \cref{tab: experiment_sota}, Face-MoGLE achieves the highest IR scores in multimodal (0.7014) and mask-to-face (0.0398) generation, and ranks second in text-to-face (0.9527), indicating that the generated images are both semantically aligned and preferred by human observers. \cref{fig: vis_results} highlights these results, where Face-MoGLE effectively resolves cross-modal conflicts and generates more natural features in multimodal and mask-to-face tasks. Although slightly behind in text-to-face IR, it achieves better geometric accuracy and visual-textual consistency, striking a balance between structural fidelity and semantic alignment. More results are available in~\cref{sec: more_results}.

\subsection{Ablation Studies}

\begin{table}[h]
\centering
\caption{Effect of harmonizing global and local experts.}
\label{tab: ablation_global_local}
\begin{tabular}{lcccc}
\toprule
\textbf{Expert Composition} & \textbf{FID}~$\downarrow$ & \textbf{KID}~$\downarrow$ & \textbf{Mask}~$\downarrow$ & \textbf{Text}~$\uparrow$ \\
\midrule
Only Global & \underline{30.36} & \underline{18.16} & \underline{2.47} & 26.30 \\
Only Local & 33.62 & 20.45 & 4.87 & \textbf{27.07} \\
\rowcolor{gray!20}Global \& Local & \textbf{22.24} & \textbf{10.87} & \textbf{2.44} & \underline{26.32} \\
\bottomrule
\end{tabular}
\end{table}

\subsubsection{Effect of Harmonizing Global and Local Experts} We conduct ablation studies on three configurations: \textit{Global Expert} (holistic semantic mask), \textit{Local Experts} (decoupled binary masks), and the \textit{Combined Global + Local Experts}, as summarized in~\cref{tab: ablation_global_local}. The Global Expert alone yields a moderate FID of 30.36 but shows poor semantic mask alignment (Mask: 2.47). In contrast, Local Experts, while achieving the best text alignment (Text: 27.07) due to precise mapping between facial regions and textual semantics, suffer from the highest FID (33.62) due to the lack of holistic spatial context. Our unified framework, which dynamically integrates global and local features via a gating network, outperforms both, achieving a significantly improved FID of 22.24 and enhanced mask consistency (2.44). These results highlight the effectiveness of hierarchical expert fusion: global experts ensure topological coherence, while local experts provide fine-grained semantic control, jointly promoting visual fidelity and structural precision.

\subsubsection{Impact of Various Gating Mechanisms.}

\begin{table}[h]
\centering
\caption{Impact of various gating mechanisms.}
\label{tab: ablation_gating_mechanisms}
\begin{tabular}{lcccc}
\toprule
\textbf{Gating Mechanism} & \textbf{FID}~$\downarrow$ & \textbf{KID}~$\downarrow$ & \textbf{Mask}~$\downarrow$ & \textbf{Text}~$\uparrow$ \\
\midrule
w/o Diffusion & \underline{25.74} & \underline{13.12} & \textbf{2.37} & \underline{26.53} \\
Scalar Gating & 43.48 & 30.58 & 4.74 & \textbf{26.72} \\
\rowcolor{gray!20}Matrix Gating & \textbf{22.24} & \textbf{10.87} & \underline{2.44} & 26.32 \\
\bottomrule
\end{tabular}
\end{table}


We evaluate three gating strategies: static weights, time-dependent scalar weights, and our spatiotemporal matrix weights. As~\cref{tab: ablation_gating_mechanisms} shows, matrix gating achieves the best FID (22.24) and KID (10.87), outperforming scalar gating by 48.9\% and static weights by 13.6\%. The severe degradation under scalar gating (FID:43.48) stems from its inability to handle spatial conflicts, whereas static weights (FID:25.74) lack temporal adaptability across diffusion stages. Our method resolves these through pixel-wise weight maps that evolve spatially and temporally, achieving optimal balance between semantic control (Mask:2.44) and photorealism. This conclusively demonstrates the necessity of spatiotemporal dynamics in expert fusion.

\subsubsection{Joint Contribution of Expert and Gating}

\begin{table}[h]
\centering
\caption{Joint contribution of expert and gating.}
\label{tab: ablation_on_components}
\begin{tabular}{cccccc}
\toprule
\textbf{Expert} & \textbf{Gating} & \textbf{FID}~$\downarrow$ & \textbf{KID}~$\downarrow$ & \textbf{Mask}~$\downarrow$ & \textbf{Text}~$\uparrow$ \\
\midrule
$\times$ & $\times$ & 33.25 & 23.07 & \underline{2.49} & \textbf{26.71} \\
$\checkmark$ & $\times$ & \underline{26.55} & \underline{14.87} & 3.20 & 26.38 \\
$\times$ & $\checkmark$ & 31.30 & 19.11 & 2.60 & \underline{26.64} \\
\rowcolor{gray!20}$\checkmark$ & $\checkmark$ & \textbf{22.24} & \textbf{10.87} & \textbf{2.44} & 26.32 \\
\bottomrule
\end{tabular}
\end{table}


We ablate the individual and joint effects of global-local experts and dynamic gating. As shown in~\cref{tab: ablation_on_components}, using only experts (FID:26.55) or gating (FID:31.30) yields partial improvements over the baseline (FID:33.25), while their combined use achieves optimal FID (22.24). This synergy arises because experts decompose facial semantics into multiple binary components (mask error drops from 3.20 to 2.44), while the gating network dynamically aligns these components across space and time. Although the baseline shows marginally higher text alignment (Text:26.71 vs. 26.32), its overly smoothed outputs lack semantic precision, whereas our full model balances photorealism and control. This validates that global-local experts and adaptive gating are mutually essential for high-fidelity and controllable generation.

\subsubsection{Zero-Shot Generalization Validation}
As shown in \cref{tab: experiment_zero_sota}, Face-MoGLE achieves superior zero-shot generalization on the MM-FFHQ-Female dataset, outperforming existing methods on most metrics. Our framework attains state-of-the-art image quality (FID: 62.93, KID: 31.27), mask consistency (Mask: 2.77), text alignment (CLIP: 28.06), and human preference (Image Reward: 0.1801). While TediGAN shows better CMMD performance (1.091 vs. ours 1.238), this can be attributed to its StyleGAN backbone being pre-trained on the FFHQ dataset. Compared to the best diffusion-based baseline (UaC), our method reduces FID by \textbf{27.3\%} and KID by \textbf{32.9\%}, demonstrating the effectiveness of our mask-decoupling strategy and global-local MoE architecture. The dynamic gating network enables adaptive feature fusion during the diffusion process, contributing to robust performance on unseen semantic combinations. These results validate Face-MoGLE's capability to synthesize high-fidelity faces under zero-shot conditions without task-specific fine-tuning.

\subsubsection{Evaluation with Deepfake Detection}
We evaluate the ability of our synthesized faces to evade deepfake detectors, as shown in \cref{tab: experiment_on_deepfake_detection}. Specifically, we test against \textit{NPR} \cite{npr}, a general-purpose detector, and \textit{Wavelet-CLIP} \cite{wavelet-clip}, which focuses on face-specific artifacts. Face-MoGLE achieves near-random AUC on NPR (0.50 vs. Collaborative's 0.51) and significantly outperforms Collaborative on Wavelet-CLIP (0.46 vs. 0.75). In light of these results, we stress that this evaluation is conducted solely for defensive research purposes. Societal impacts and responsible AI are discussed in~\cref{sec: ethics}.


\begin{table}[h!]
\centering
\caption{Performance comparison of two deepfake detection models against different face generation methods. Each cell is shown as NPR~\cite{npr} / Wavelet-CLIP~\cite{wavelet-clip} detection results.}
\label{tab: experiment_on_deepfake_detection}
\begin{tabular}{lccc}
\toprule
\textbf{Method} & \textbf{AUC} & \textbf{EER} & \textbf{AP} \\
\midrule
TediGan~\cite{tedigan} & 0.73 / 0.81 & 0.35 / 0.26 & 0.71 / \underline{0.76} \\
UaC~\cite{uniteandconquer} & 0.45 / 0.96 & 0.53 / 0.11 & 0.43 / 0.96 \\
Collaborative~\cite{collaborative} & \underline{0.51} / \underline{0.75} & \underline{0.51} / \underline{0.32} & \underline{0.50} / 0.79 \\
PixelFace+~\cite{pixelfaceplus} & 0.28 / 0.87 & 0.65 / 0.20 & \textbf{0.37} / 0.89 \\
\rowcolor{gray!20}Face-MoGLE (Ours) & \textbf{0.50} / \textbf{0.46} & \textbf{0.50} / \textbf{0.53} & 0.52 / \textbf{0.46} \\
\bottomrule
\end{tabular}
\end{table}

\section{Conclusion}
In this work, we introduced a mixture of global and local experts with a diffusion transformer for controllable face generation. Our method effectively decouples semantic mask information and dynamically selects experts to enhance image synthesis fidelity and condition alignment. Through extensive experiments, we demonstrated improvements in multimodal and monomodal generation and robust generalization. Future work may explore more efficient architectures and applications in real-world scenarios.

\bibliographystyle{ACM-Reference-Format}
\bibliography{refs}


\appendix

\section{Societal Impacts and Responsible AI}~\label{sec: ethics}

Our research focuses on controllable face generation, based on a diffusion transformer architecture combined with a mixture of global and local experts, aiming to support a variety of optimistic application scenarios. This technology is not intended to mislead or deceive. However, similar to other generative models, it may still be misused for impersonating individuals. We strongly oppose any behavior that produces deceptive or harmful facial content.

While acknowledging the potential risks of misuse, we also recognize the significant positive potential of this technology. Our method can be broadly applied in digital creativity, virtual human interaction, and personalized content generation. Additionally, it holds promise in public-interest applications such as generating portraits of missing children or criminal suspects (e.g., by reconstructing facial contours from the semantic mask and inferring other attributes from textual descriptions), thereby contributing to public safety and social welfare. We are committed to the responsible development of AI technologies that benefit humanity.

To mitigate potential misuse and provide necessary safeguards, we are also exploring the application of our method in advancing face forgery detection. Specifically, we use the generated facial images as training data to support the development of general-purpose face forgery detection models. Preliminary experiments show that incorporating data generated by our method improves the generalization ability of these models. We will continue to share our latest progress with the research community actively.

\section{Dataset Details}~\label{sec: dataset details}

\subsection{MM-CelebA-HQ}
This dataset contains 30{,}000 high-resolution facial images, each annotated with a corresponding semantic segmentation map and ten natural language descriptions. The semantic segmentation maps label each pixel into one of 19 categories, including \textit{background}, \textit{face skin}, \textit{nose}, \textit{eyeglasses}, \textit{left eye}, \textit{right eye}, \textit{left eyebrow}, \textit{right eyebrow}, \textit{left ear}, \textit{right ear}, \textit{inner mouth}, \textit{upper lip}, \textit{lower lip}, \textit{hair}, \textit{hat}, \textit{earring}, \textit{necklace}, \textit{neck}, and \textit{clothing}. These segmentation labels provide pixel-level structural and contextual information useful for supervised learning and evaluation in image generation tasks. Each image is also paired with 10 unique text descriptions that capture detailed visual characteristics, such as facial features, expressions, accessories, age, and gender. During training, one of the 10 descriptions is randomly selected, while during testing, the first description is always used to ensure consistency. The dataset serves as a comprehensive benchmark for text-to-image generation and multimodal learning, supporting tasks such as conditional image synthesis, semantic-guided generation, and multimodal learning.

\subsection{MM-FFHQ-Female} 

FFHQ-Text~\cite{ffhq_text} is a smaller-scale but highly specialized dataset that comprises 760 high-quality female face images from the FFHQ (Flickr-Faces-HQ)~\cite{ffhq} dataset. Each image is paired with 9 distinct natural language descriptions, detailing fine-grained facial attributes such as makeup style, hairstyle, facial expression, skin tone, and accessories. These descriptions emphasize subtle details and variations, making the dataset particularly suitable for evaluating text-to-image generation and manipulation tasks that require high sensitivity to nuanced text cues. To produce semantic masks for FFHQ-Text, we leverage two pretrained facial parsing models—FaRL~\cite{farl} and SegFace~\cite{segface}. For masks with overall accuracy (OA) below 0.8, we conduct manual annotation, whereas for those achieving OA $\geq$ 0.8, we adopt a randomized sampling strategy, comprising 90\% from FaRL and 10\% from SegFace. A randomly sampled textual description for each image is used during zero-shot evaluation to ensure consistency across experiments. The combination of detailed textual annotations, semantic masks, and high-resolution facial images enables comprehensive studies on fine-level semantic alignment and learning in multimodal models. We will release this dataset to promote community development.



\section{More Results}~\label{sec: more_results}

\subsection{Multimodal Face Generation}
\begin{figure*}[t]
    \centering
    \includegraphics[width=0.9\linewidth]{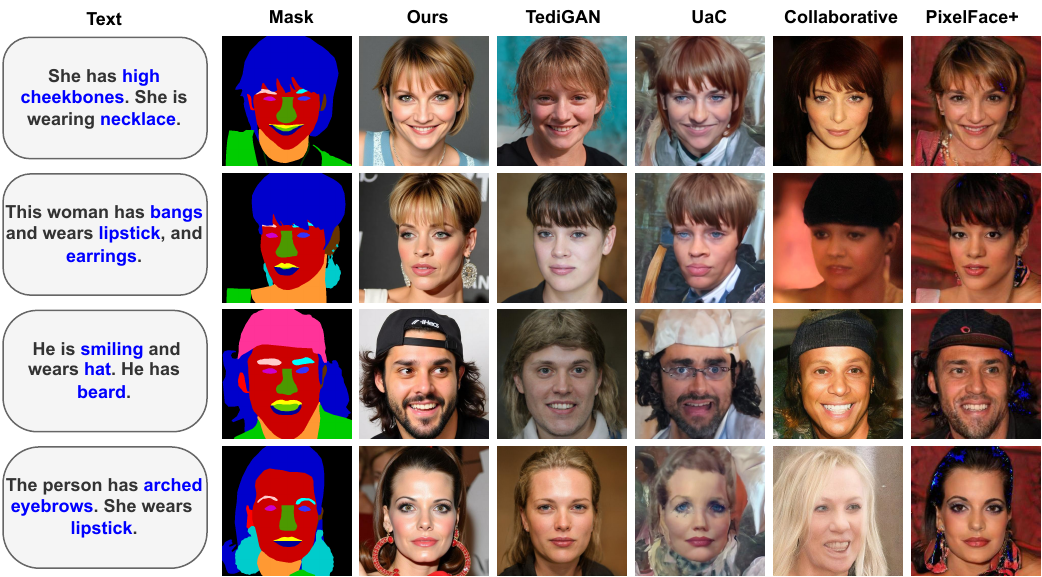}
    \caption{Comparative results of multimodal face generation on the MM-CelebA-HQ dataset.}
    \label{fig: more_multimodal_generation}
\end{figure*}

\cref{fig: more_multimodal_generation} demonstrates the comparative results of multimodal face generation between our method and several state-of-the-art multimodal generation methods. As shown, our method consistently produces more realistic and semantically faithful faces, effectively integrating multiple modalities such as text descriptions and segmentation masks. Noticeably, our generated faces exhibit finer facial details and more accurate modality alignment than other methods.

\subsection{Mask-to-Face Generation}
\begin{figure*}[t]
    \centering
    \includegraphics[width=0.9\linewidth]{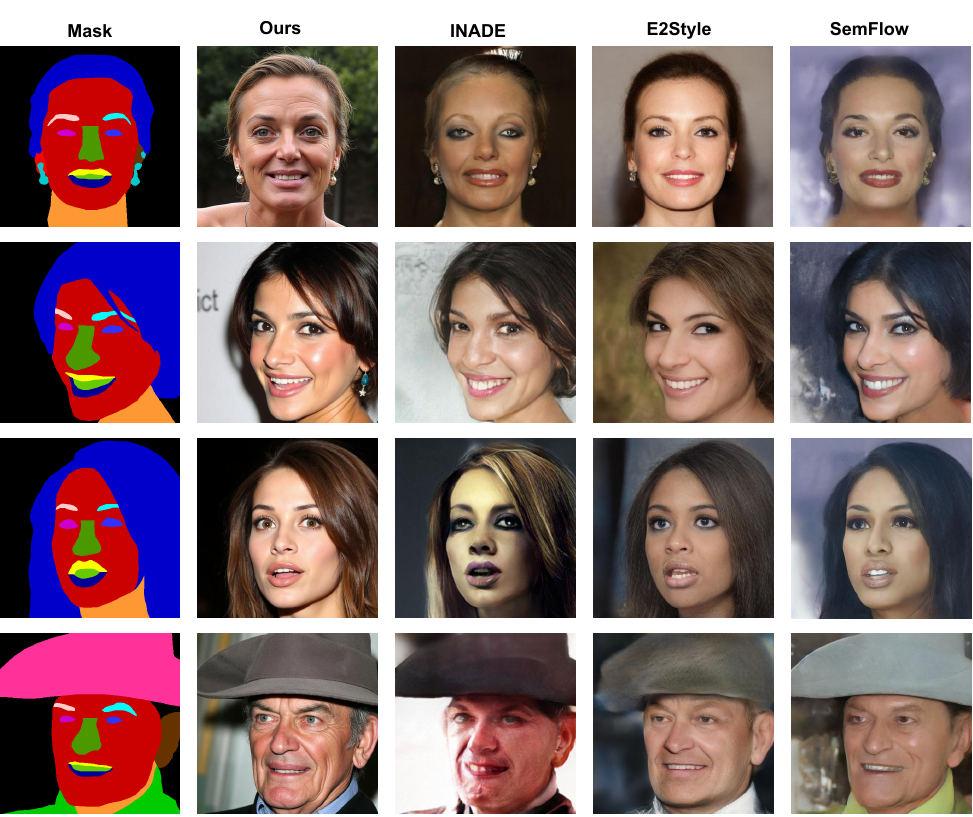}
    \caption{Comparative results of mask-to-face generation on the MM-CelebA-HQ dataset.}
    \label{fig: more_mask_to_face_generation}
\end{figure*}

In~\cref{fig: more_mask_to_face_generation}, we present a comprehensive comparative visualization of mask-to-face generation performance. Although our method is not explicitly designed for the mask-to-face generation task, it still achieves commendable structural alignment with the input semantic masks, yielding compelling images of superior fidelity.

\subsection{Text-to-Face Generation}
\begin{figure*}[t]
    \centering
    \includegraphics[width=0.9\linewidth]{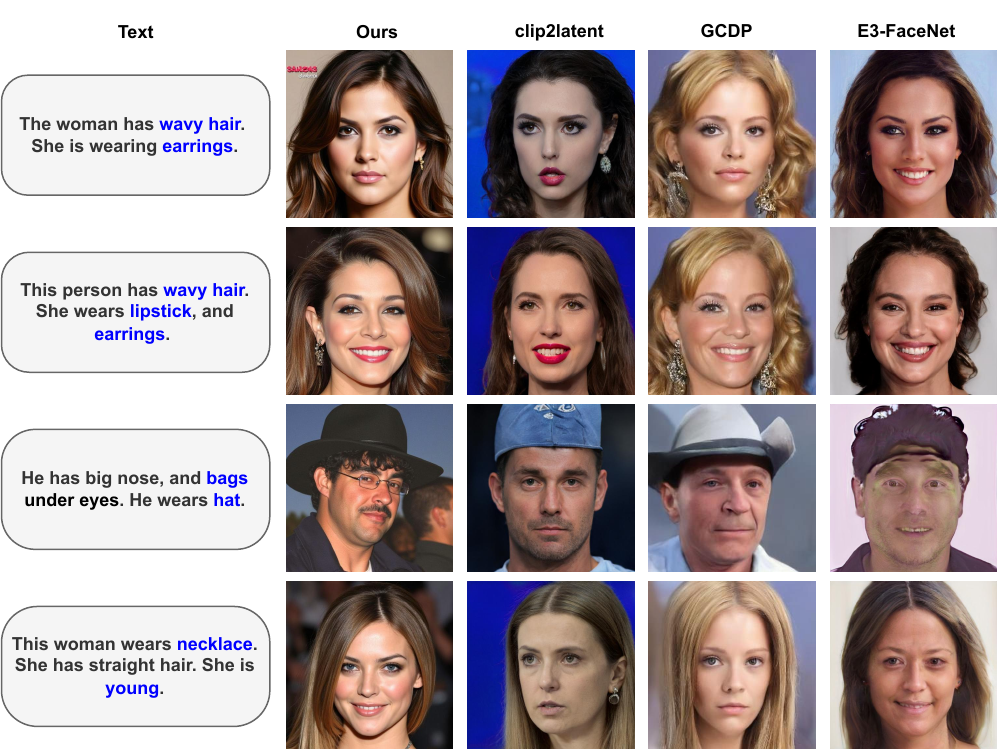}
    \caption{Comparative results of text-to-face generation on the MM-CelebA-HQ dataset.}
    \label{fig: more_text_to_face_generation}
\end{figure*}

\cref{fig: more_text_to_face_generation} illustrates the comparative results of text-to-face generation methods. It is evident from the examples provided that our method surpasses previous techniques in capturing subtle textual cues and translating them accurately into visual facial features. Compared to other methods, our generated faces better reflect the described attributes, demonstrating a notable improvement in text consistency.

\subsection{Zero-shot Generalization}

\begin{figure*}[t]
    \centering
    \includegraphics[width=0.9\linewidth]{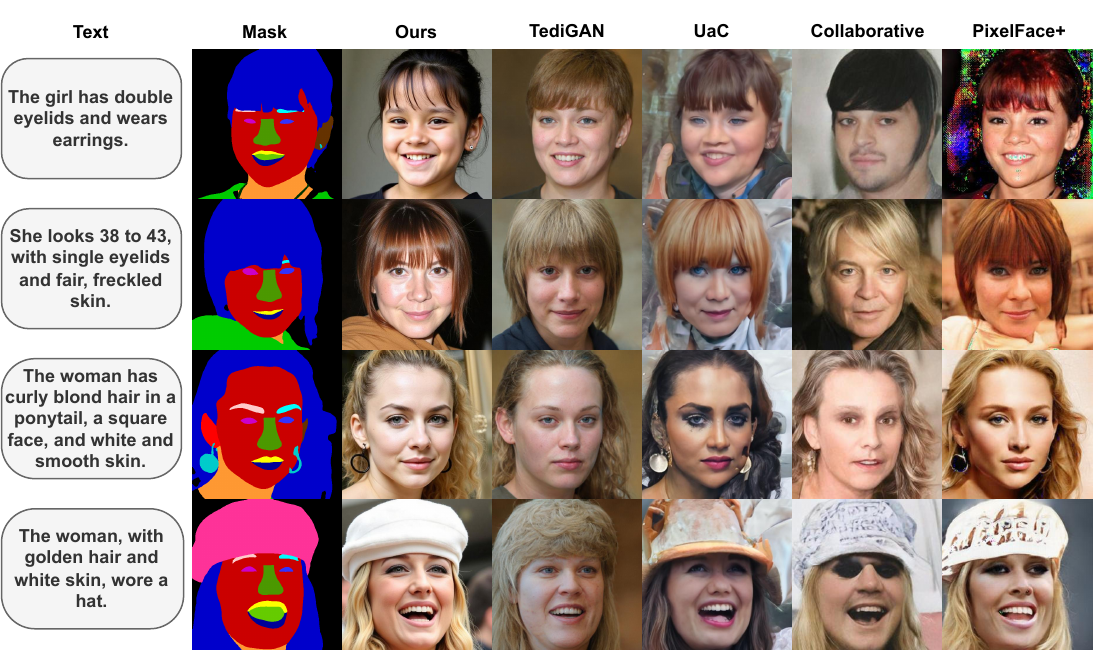}
    \caption{Comparative results of zero-shot generalization on the MM-FFHQ-Female dataset.}
    \label{fig: zero_ffhq}
\end{figure*}

\noindent\textbf{MM-FFHQ-Female}. As shown in \cref{fig: zero_ffhq}, our method produces significantly more faithful and realistic generations than baseline methods. The results demonstrate consistency with the input prompts while preserving fine-grained semantic structures with the mask.

\subsection{Ablation Studies}

\begin{figure*}[t]
    \centering
    \includegraphics[width=0.9\linewidth]{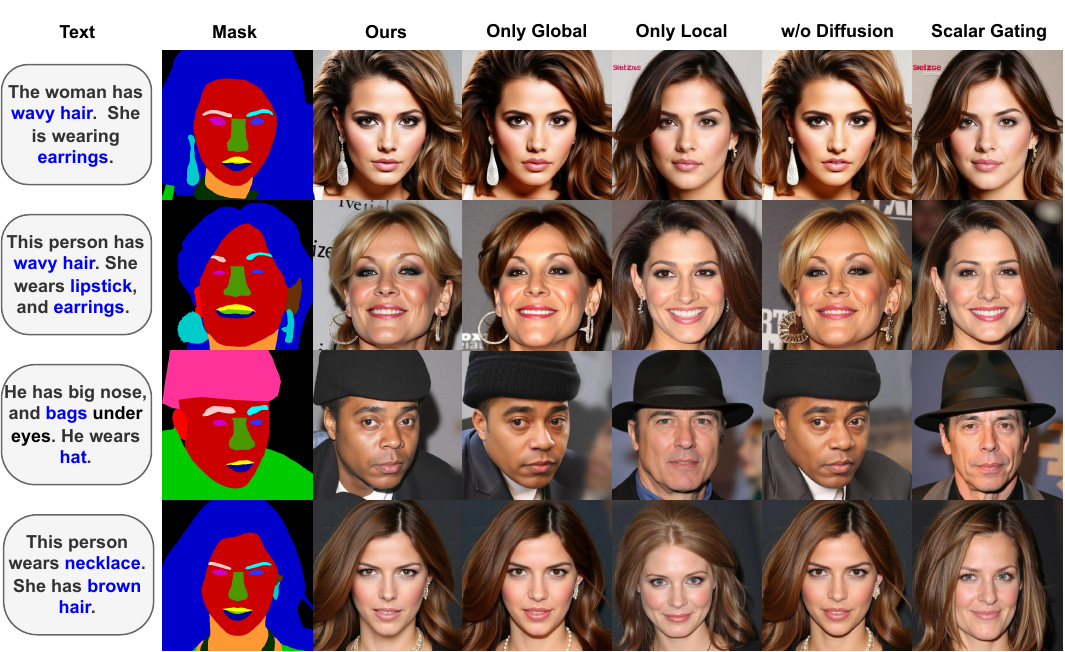}
    \caption{Comparative results of ablation studies on the MM-CelebA-HQ dataset.}
    \label{fig: more_ablation}
\end{figure*}

\cref{fig: more_ablation} illustrates the visual results of our ablation studies. The figure includes variations such as \textit{Only Global}, \textit{Only Local}, \textit{w/o Diffusion}, \textit{Scalar Gating}, and our final best-performing method.

\end{document}